\RequirePackage{fix-cm}
\documentclass[twocolumn]{svjour3}          
\smartqed  
\usepackage{graphicx}
\usepackage[comma,square]{natbib}
\usepackage{multirow}
\usepackage[table]{xcolor}
\usepackage{dirtytalk}
\usepackage{footnote}
\usepackage{float}
\usepackage{svg}
\usepackage{tikz}
\usepackage{pgfplots}
\usepackage{amsmath}
\usepackage{algorithm}
\usepackage[noend]{algpseudocode}
\usepackage{array}
\usepackage{booktabs}
\usepackage{makecell}
\usepackage{pgfplots}
\usepackage{import}
\algnewcommand\AND{\textbf{ and }}
\algnewcommand\OR{\textbf{ or }}
\algnewcommand\NOT{\textbf{ not }}
\algnewcommand\BREAK{\textbf{ break }}
\algnewcommand\RETURN{\textbf{ return }}

\begin{document}

\title{Surrogate Assisted Strategies
}
\subtitle{
The Parameterisation of an Infectious Disease Agent-Based Model
}


\author{ Rylan Perumal\textsuperscript{1} \and
         Terence L van Zyl\textsuperscript{2}
}


\institute{R. Perumal\textsuperscript{1} \at
              School of Computer Science and Applied Mathematics \\
              University of the Witwatersrand, South Africa \\
              \email{rylan.perumal@wits.ac.za}           
           \and
           T.L. van Zyl\textsuperscript{2} \at
              Institute for Intelligent Systems\\ University of Johannesburg, 
              South Africa \\
              \email{tvanzyl@uj.ac.za}
}

\date{Received: date / Accepted: date}
\newcommand\copyrighttext{%
  \footnotesize \textbf{Under Review}. Personal use of this material is permitted.  Permission from the authors must be obtained for all other uses, in any current or future media, including reprinting/republishing this material for advertising or promotional
  purposes, creating new collective works, for resale or redistribution to servers or lists, or reuse of any copyrighted component of this work in other works.}
\newcommand\arxivcopyrightnotice{%
\begin{tikzpicture}[remember picture,overlay]
\node[anchor=south,yshift=10pt] at (current page.south) {\fbox{\parbox{\dimexpr\textwidth-\fboxsep-\fboxrule\relax}{\copyrighttext}}};
\end{tikzpicture}%
}
\def\BibTeX{{\rm B\kern-.05em{\sc i\kern-.025em b}\kern-.08em
    T\kern-.1667em\lower.7ex\hbox{E}\kern-.125emX}}

\newcommand{\code}{\texttt}
\newcolumntype{P}[1]{>{\centering\arraybackslash}m{#1}}
\maketitle
\arxivcopyrightnotice
\begin{abstract}
Parameter calibration is a significant challenge in agent-based modelling and simulation (ABMS). An agent-based model's (ABM) complexity grows as the number of parameters required to be calibrated increases. This parameter expansion leads to the ABMS equivalent of the \say{curse of dimensionality}. In particular, infeasible computational requirements searching an infinite parameter space. We propose a more comprehensive and adaptive ABMS Framework that can effectively swap out parameterisation strategies and surrogate models to parameterise an infectious disease ABM. This framework allows us to evaluate different strategy-surrogate combinations' performance in accuracy and efficiency (speedup). We show that we achieve better than parity in accuracy across the surrogate assisted sampling strategies and the baselines. Also, we identify that the Metric Stochastic Response Surface strategy combined with the Support Vector Machine surrogate is the best overall in getting closest to the true synthetic parameters. Also, we show that DYnamic COOrdindate Search Using Response Surface Models with XGBoost as a surrogate attains in combination the highest probability of approximating a cumulative synthetic daily infection data distribution and achieves the most significant speedup with regards to our analysis. Lastly, we show in a real-world setting that DYCORS XGBoost and MSRS SVM can approximate the real world cumulative daily infection distribution with $97.12$\% and $96.75$\% similarity respectively.


\end{abstract}

\section{Introduction}
\label{intro}

Agent-based models (ABMs) offer the possibility to model many complex real-world scenarios~\citep{abm_use_case}. These scenarios range from modelling the spread of an epidemic within a population to modelling trends based on agents' behaviour in the stock market. The number of model parameters to be calibrated to match real-world data rises with increased model complexity~\citep{dengue}. As a result of the larger parameter space, searching for meaningful values can become computationally prohibitive. Machine learning (ML) models, namely surrogate models (SMs), can expedite the search of any parameter space and, in so doing, improve both the accuracy and efficiency of ABMs. SMs are used to classify whether a candidate parameterisation is a good parameter combination, allowing ABMs to match real-world data~\citep{surrogate_og}. Other uses of surrogate optimisation strategies include the global optimisation of expensive multimodal functions where derivatives are unavailable ~\citep{srbf_2007,dycors_2013}.

Expanding on previous research, we present an improved Framework to facilitate parameterisation of infectious disease ABMs in Figure \ref{fig:ImprovedFramework}. We further implement a new surrogate assisted sampling strategy and a surrogate assisted evolutionary parameterisation strategy within this framework. We modified and re-implemented the MSRS and DYCORS surrogate optimisation strategies to enable ABM parameterisation within our framework. Lastly, we select the best strategy-surrogate combinations in terms of accuracy and efficiency (speedup) and minimisation of Standardised L$_{2}$ Norm and use each to parameterise the ABM towards a real-world cumulative daily infection distribution. Our results obtained show:

\begin{enumerate}
    \item We obtains significant speedup between two to fours times above the baseline models using surrogate assisted data drive optimisation.
    \item We are able to achieve better than parity accuracy across multiple parameters using surrogate assisted strategies when compared to the baselines.
    \item The MSRS SVM strategy-surrogate combination is able to minimise the distance to the synthetic ABM parameters the best overall.
    \item The best overall method in terms of both accuracy and speedup is DYCORS XGBoost.
    \item DYCORS XGBoost and MSRS SVM used within our ABMS Framework when tested on real-world cumulative daily infection distribution for South Africa are able to get approximately $97.12$\% and $96.75$\% accuracy respectively.
\end{enumerate}

\section{Background and Related Work}
\subsection{Agent-Based Modelling and Simulation for Infectious Disease Epidemiology}

Agent-based modelling and simulation (ABMS) is an effective natural fit for modelling infectious diseases. Agent-based models (ABMs) are capable of modelling interactions between individuals and their environment. Further, they can capture unexpected emergent patterns and trends during an epidemic that result from collective individual agent behaviours and interactions \citep{taxonomy}. Each agent within an ABM can have different characteristics, more closely representing the variation in human populations. The agents act autonomously, governed by the interaction of their set of pre-defined rules and distinctive characteristics. This autonomy allows ABMs to simulate many complex real-world scenarios with sufficient fidelity~\citep{abm_use_case}. ABMS can also be used as a substitute for a real-world epidemiological study since it can often be infeasible or even impossible to run a real-world experiment~\citep{abm_methods_techniques, public_health_abm}. 

While there are significant benefits to using ABMs for infectious disease epidemiology, there are equally limitations. ABMs ordinarily require long run times due to the increased computational complexity resulting from agent interactions incorporated into the model \citep{surrogate_improve, surrogate_og}. Additionally, model validation and parameterisation present significant challenges in ABMS, precisely when matching real-world data~\citep{dengue}. Of these two challenges, we are particularly interested in the parameterisation of ABMs. Difficulty finding correct parameter combinations of ABMs leads to extensive calibration efforts resulting in increased model development time. As more complexity is added to the model, the parameter space expands, leading to the ABMS equivalent of the \say{curse of dimensionality} problem. The outcome is impractical memory and computational costs when searching for meaningful parameter combinations~\citep{bayesian_abm, influenza_review}.

\subsection{Surrogate Models in Agent-Based Modelling and Simulation}

One approach to overcoming the computational limitations of ABMS is the use of surrogate models (SMs). SMs are machine learning (ML) models that act as function approximators to an agent-based model. Additionally, SMs provide a computationally tractable solution to addressing parameter sensitivity analysis, robust analysis and empirical validation in ABMS~\citep{van_deHoog_surrogate}. These properties make SMs appealing for approximating complex ABMs that are computationally expensive to validate and calibrate. Previously, Gaussian process regression, also known as the Kriging method, has been used as a surrogate modelling approach to facilitate parameter space exploration and sensitivity analysis challenges in ABMS \citep{surrogate_og, surrogate_improve}. However, Kriging's performance is dependent on the model's ability to estimate the spatial continuity of the data~\citep{bargigli_kriging, dosi_kriging}.

\cite{surrogate_og} presented an alternate approach, overcoming some of the limitations of the Kriging method. An iterative algorithm is proposed for training a SM to effectively approximate the ABM. The novel approach is realised by combining ML and intelligent iterative sampling. It is demonstrated that a model's parameter space can be effectively searched utilising fewer computational resources adopting their approach. In their work the XGBoost ML algorithm is used, where the SM is built in a stage-wise fashion, allowing optimisation of an arbitrary differentiable loss function. This method has also been applied to an Asset Pricing Model by \cite{asset_pricing} and a Island Growth model by \cite{island_growth}. The results obtained show that the SM is an accurate function approximator of the ABM. Further, the SM radically reduced the computation time for large-scale parameter space calibration and exploration. \cite{surrogate_improve} improve on the work of \cite{surrogate_og} replacing XGBoost with the CatBoost algorithm. They show the surrogate can better approximate the ABM and as a result further reduce parameter calibration and exploration time.

\subsection{Surrogate Assisted Optimisation}
Many real-world optimisation problems involve high-dimensional black-box functions that are outputs of computationally expensive simulations. Generally, finding the global optimum of these problems is unrealistic as it requires a significant amount of function evaluations~\citep{dycors_2013}. It is often the case that the derivatives of these black-box functions aren't available. Therefore, derivative-free strategies have been developed to allow for their optimisation. In particular, we are interested in derivative-free optimisation and derivative-free heuristic methods. A common approach to derivative-free optimisation is the use of surrogate models (SMs) or metamodels. \cite{srbf_2007} present a method for the global optimisation of expensive multimodal functions. They use a response surface model (radial basis function and neural network) as a SM. The method iteratively uses the surrogate to approximate the output of the expensive multimodal function and then selects the best potential candidate. The candidate is selected based on two criteria: the estimated response from the SM and the minimum distance to the previously evaluated points. The results presented indicate that this method is a promising solution for the optimisation of expensive high dimensional problems. \cite{dycors_2013} combines a radial basis function SMs and dynamic coordinate search for the global optimisation of computationally expensive functions. This is an extension of the previous work presented by \cite{srbf_2007}. They present two algorithms where they show that their work is an improvement on classical approaches, especially for high dimensional problems. These surrogate optimisation approaches are appealing as they provide an abstraction towards addressing the parameterisation challenges in ABMs. Specifically, in epidemiology, the accurate and efficient parameterisation of infectious disease models is imperative. 


\section{Agent-Based Modelling and Simulation Framework}
\begin{figure}[!htb]
    \centering
    \includegraphics[scale=0.23]{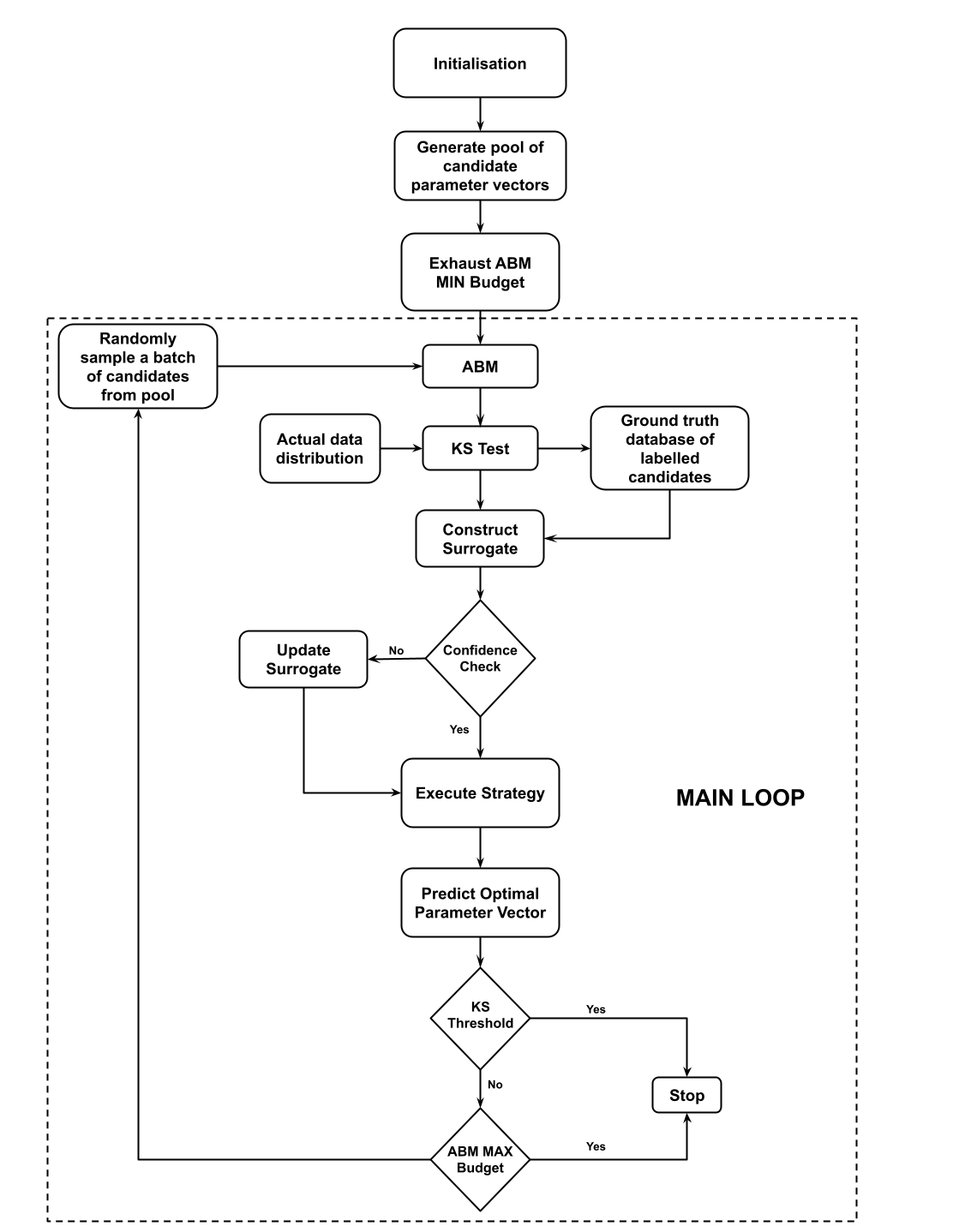}
    \caption{Agent-Based Modelling and Simulation (ABMS) Framework}
    \label{fig:ImprovedFramework}
\end{figure}

Figure \ref{fig:ImprovedFramework} represents the improved ABMS framework inspired by the algorithm of \cite{surrogate_og} and \cite{perumal2020}. Our framework starts of by setting an initial configuration for the parameterisation task. The details of the initial configuration is as follows:
\begin{itemize}
    \item selection of a sampling method, which will be used generate candidate parameters from the parameter space;
    \item selection of a machine learning algorithm that will construct the SM;
    \item selection of a paramterisation strategy;
    \item actual (real/synthetic) data distribution we are parameterising towards;
    \item setting the \textit{ABM MIN/MAX Budget} limits which represent the minimum /maximum number of samples required to be evaluated by the agent-based model;
    \item defining an initial level of significance for the Kolmogorov-Smirnov Test;
    \item lastly, we define a confidence criteria for a given SM.
\end{itemize}
Our proposed ABMS Framework can integrate different surrogates, parameterisation strategies and sampling methods. After initialisation, a pool of candidate parameter vectors is generated utilising the selected sampling method. During initialisation, we exhaust the \textit{ABM MIN Budget}. We sample a subset of candidates, equal to \textit{ABM MIN Budget}, from the parameter pool, which the ABM then evaluates. The ABM generates a simulated data distribution based on the input candidate. We compare the similarity between the actual data distribution and the simulated data distribution using Equation \ref{equation1} and the corresponding candidate is labelled accordingly. The labelled candidates are then included in the ground-truth database. We then construct the SM employing the ground-truth database at the first iteration of the \textit{Main Loop} . After the SM has been constructed, we execute the strategy. If we are not at the first iteration, we check whether the SM has diverged from the \textit{Confidence Criteria}. Depending on whether we have diverged from the \textit{Confidence Criteria}, we either go straight to execute the strategy or update the SM using the newly evaluated batch of candidates. Once the strategy has been executed, we predict the optimal candidate. Depending on whether the \textit{Kolmogorov-Smirnov Threshold} (similarity threshold between two distributions) or \textit{ABM MAX Budget} is met, we either stop, or we go into the next iteration of the framework. During the next iteration, a batch of candidates are randomly sampled from the pool, and we continue within the \textit{Main Loop}.

\subsection{Infectious Disease Agent-Based Model}
The agent-based model (ABM) used in this framework is a pre-existing model. The ABM used is a continuous space virus spread model, where the disease transmission dynamics follows the basic Susceptible-Infected-Recovered (SIR) framework, proposed by Kermack and McKendrick \cite{kermack1927_sir}, simulated for $41$ days. The SIR framework models the ratio of susceptible, infected and recovered individuals within a population. Also, once infected individuals become aware that they are infected (i.e. they have detected the infection), they are then made immovable to represent being in a state of quarantine (lockdown). The ABM takes the parameters presented in Table \ref{table1} as input, simulates an epidemic based on the input and then generates an infection data distribution based on the simulation. 


\begin{table}[htb!]
\centering
\caption{
Table of the ranges for each parameter value of the ABM that we have considered for parameterisation. Parameters $1, 2, 3,$ and $6$ are sampled between the range $(0, 1)$. Parameter $7$ is sampled between the range $(0, 0.022)$ to mimic real world interaction as the space defined is bounded by $(1, 1)$. Parameters $4$ and $5$ are sampled between the range $(0, 41)$ days. 
}
\begin{tabular}{l|l}
\toprule
\multicolumn{1}{c}{\textbf{Parameters}} & \multicolumn{1}{|c}{\textbf{Range}} \\
\bottomrule \toprule
1. Transmission Probability ($\beta$) & (0.0, 1.0) \\ 
2. Reinfection Probability & (0.0, 1.0) \\ 
3. Death Probability & (0.0, 1.0) \\ 
4. Infection Period & (0, 41) days  \\ 
5. Detection Time & (0, 41) days \\ 
6. Speed & (0.0, 1.0) \\ 
7. Interaction Radius & (0.0, 0.022)  \\ 
\bottomrule
\end{tabular}
\label{table1}
\end{table}

\subsection{Kolmogorov-Smirnov Test}
The two-sample Kolmogorov-Smirnov (KS) test is used to compare the similarity between the distributions of the actual and simulated data as follows:
\begin{equation}
    D_{A, S} = \sup_{x} \large| F_{A}(x) - F_{S}(x) \large|, \label{equation1}
\end{equation}
where $x$ represents the feature we are measuring (number of infected individuals) and $F_{A}$ and $F_{S}$ are the distribution functions of the actual and simulated data respectively. The null hypothesis is $H_{0}:$ the two distributions are not the same. The null hypothesis is rejected at a significance level $\alpha = 0.05$ if $D_{A, S} > D_{N, \alpha}$, where 
\begin{equation}
    D_{N, \alpha} = c(\alpha)\sqrt{\frac{2 \cdot N}{N^2}},
\end{equation}
\begin{equation}
    c(\alpha) = \sqrt{-\ln{(\alpha)} \cdot \Big(\frac{1}{2}\Big)}
\end{equation}
and $N=$ Population Size. 
A candidate parameter vector generates the simulated data. The vector is labelled as negative if the two distributions are not the same and positive if they are similar. The KS test compares the cumulative distributions of the samples, which must be calculated. The data distributions that we are comparing are time series. However, the empirical KS test is formulated to assess the distance between two independent and identically distributed samples. To use the KS test on our time-series data, we convert the time-series to a cumulative distribution function by first performing a cumulative sum along the time dimension. We then scale the cumulative sum to arrive at a cumulative distribution that maintains the time series's integrity. This normalisation makes our problem scale-invariant to $N$, which allows us to measure the similarity of the exact epidemic trends between the two scale-invariant time-series distributions.

\subsection{Sanity Check}
To ensure that our implemented ABMS framework, seen in Figure \ref{fig:ImprovedFramework}, is reliable, we conduct a sanity check. Given a set of known optimal parameters, $\theta^{*}$, we use this as input to the constructed ABM. The ABM then generates a synthetic data distribution based on the known parameters, $\theta^{*}$, as output. We subsequently use the synthetic distribution as the actual data distribution in Figure \ref{fig:ImprovedFramework}. After that, we execute the ABMS Framework and observe if we can approximate known $\theta^{*}$. We use different parameter combinations of $\theta^{*}$ within our framework to assess overall generalisation.

\section{Surrogate Assisted Parameterisation Strategies}

The surrogate assisted optimisation methods, Metric Stochastic Response Surface (MSRS) and DYnamic COOrdinate Search Using Response Surface Models (DYCORS) has been modified from the literature. We have incorporated our ABM evaluation and added additional functionality that allows integration with our ABMS Framework. In addition, we implement a new surrogate assisted sampling method and a surrogate assisted evolutionary strategy for the parameterisation of the infectious disease agent-based model.  

\subsection{Surrogate Assisted Optimisation Strategies}
\subsubsection{Metric Stochastic Response Surface}
The initial lower and upper bounds of the parameters are reassigned such that each new candidate is generated as a \textit{Normal}(0, $\sigma^2$) distributed perturbation around the current best candidate. The value of $\sigma$ is increased if we have an accurate surrogate model (SM) and decrease if it is inaccurate. A sample of candidates is generated using a random sampling method. The SM predicts the Kolmogorov-Smirnov Test Statistic (KSTS) value of the newly generated candidate points and then computes the previously evaluated candidates' distance. The predicted KSTS value represents the expected similarity between the actual data distribution and simulated distribution as if the candidate was evaluated by the agent-based model (ABM). The SM's predictions and the distance to the previously evaluated points are computed and then rescaled through a linear transform on the interval $[0, 1]$. A candidate that minimises the weight-distance merit function,
\begin{equation}
  \operatorname{merit}(\Bar{x}) = w s(\Bar{x}) + (1-w) d(\Bar{x}), \label{equation2}  
\end{equation}
is selected as the optimal candidate, where $s(x)$ is the SM's prediction of candidate $x$, $d(x)$ is the minimum distance to a previously seen candidates and $0 \leq w \leq 1$. The weight $w$ is commonly cycled through a finite set of values in order to encourage exploration and exploitation, we chose $w \in \{0.3, 0.5, 0.7, 0.95\}$. When $w$ is close to $0$, we do exploration, while $w$ close to $1$ does exploitation. At the end of each iteration, the predicted best candidate is evaluated using the ABM, and that candidate is added to the ground truth database. 

\subsubsection{Dynamic Coordinate Search Using Response Surface Models}
Dynamic Coordinate Search Using Response Surface Models (DYCORS) is a modification of the Metric Stochastic Response Surface (MSRS) method presented by \cite{srbf_2007}. This method incorporates an idea from the dynamically dimensioned search (DDS) strategy presented by \cite{tolson2007dynamically}. DDS is a heuristic method for box-constrained optimisation, which scales the search for global solutions based on the maximum number of evaluations. In each iteration, a set of new candidate vectors are created by adding random perturbations of the current best solution $x_{best}$, where $x_{best}$ is a candidate vector with the lowest Kolmogorov-Smirnov Test Statistic (KSTS) value. We probabilistically determine a subset of values to perturb where the perturbations are normally distributed with mean zero and a fixed standard deviation within each candidate vector. In this method, we use DDS to create a new set of candidates perturbed at the current best solution. A SM is then used to predict the KSTS value for each of the new candidates. The candidate with the lowest predicted KSTS value is selected and evaluated by the agent-based model, and then the algorithm iterates.




\subsection{Surrogate Assisted Sampling Strategies}


\subsubsection{Surrogate Assisted Random Sampler}

A SM is trained and validated with $3$-fold cross-validation. We use the ground truth database of evaluated parameter combinations for this task where the best surrogate is selected based on the F$1$-Score. If the best surrogate's F$1$-Score is greater than or equal to a specified threshold and the number of positive and negative samples seen in the database is greater than or equal to $\phi$, where 
\begin{equation}
    \phi = (n_{folds} + n_{parameters}) + 1,
\end{equation}
we are then able to proceed with the strategy. A temporary pool of candidate parameters is generated and then classified by the surrogate as positive or negative parameter calibrations compared to the input data distribution. The new candidate pool is then generated using an $\epsilon$-greedy algorithm, where $\epsilon=0.10$. We select positively predicted candidates at a rate $1-\epsilon$ and select negatively predicted candidates at a rate $\epsilon$ from the temporary pool. The purpose of the $\epsilon$-greedy algorithm is to encourage exploration whilst still maximising the parameter space's exploitation. As a baseline for this strategy, we use a Random Sampler which generates arbitrary candidates per the ranges presented in Table \ref{table1}.

\subsubsection{Surrogate Assisted Quasi-Random Sobol Sampler}
The Quasi-Random Sobol sampling method, proposed by \cite{sobol}, generates low-discrepancy-sequences of equally distributed points on a $n$-dimensional hypercube, where $n$ refers to the number of parameters we are parameterising. We utilise this sampling method to generate a pool of candidate parameter vectors. Like the surrogate assisted random sampling strategy, we classify the candidate points as positive or negative and then use an $\epsilon$-greedy algorithm to generate a new candidate pool. As a baseline for this strategy, we use a standard Quasi-Random Sobol sampling approach.

\subsection{Covariance Matrix Adaptation Evolutionary Strategy}
We introduce a new surrogate assisted parameterisation strategy built upon the Covariance Matrix Adaptation Evolutionary Strategy (($\mu/\mu,\lambda$)-CMA-ES). We select the parent pool size, $\lambda$,  to be equal to our Batch Size size. We follow the standard algorithm using the SM to evaluate the fitness for each of the offspring. Once the new elite parents have been selected, we isolate them for our batch to be evaluated by the ABM. The mean of the next generation is calculated using the elite population. The next generation's covariance matrix is calculated using the elite population along with the mean value of the entire population at the current generation. A new set of candidate points are sampled using a Gaussian distribution with the mean and covariance of the next generation~\citep{cmaes_hasen_1996}.

\subsection{Surrogate Models}
We evaluated the following machine learning algorithms for learning surrogate models (SMs):
\begin{itemize}
    \item \textbf{eXtreme Gradient Boosting (XGBoost):} A decision tree ensemble machine learning algorithm, based off the framework by \cite{friedman}, that is scalable and efficient in its implementation \citep{xgboost}.
    \item \textbf{Decision Tree (DT):} A classification algorithm based off a tree-like structure, where the leaves represent a feature/attribute and the branches represent the decision rule which leads to the outcome of that decision \citep{dt}. 
    \item \textbf{Support Vector Machine (SVM):} A classification algorithm which finds a hyperplane in an $n$-dimensional space in order to differentiate between different classes of data points \citep{svm}.
\end{itemize}
We use the newly sampled batch of labelled parameter vectors at each iteration of the ABMS Framework to validate the SM's performance. In order to validate a SM for classification, we use the: 
\begin{equation}
    \mbox{F1 Score} = 2\cdot\frac{precision \cdot recall}{precision + recall} \label{equation4}.
\end{equation}

In order to validate a SM predicting a real value we use the:
\begin{equation}
    \mbox{RMSE} = \sqrt{\frac{\sum_{i=1}^{B}(y_{i} - \hat{y_{i}})^{2}}{B}} \label{equation5},
\end{equation}
where $y_{i}$ is the predicted real value of the $i$th candidate from the newly evaluated batch, $\hat{y_{i}}$ is the true value of the $i$th candidate and $B=$ Batch Size.

\subsection{Experiment Setup}
The following initial configurations are set as: \textit{ABM MIN Budget} $= 500$, \textit{ABM MAX Budget} $= 2500$, \textit{Batch Size} $= 250$, \textit{KS Threshold} $= 0.005$ ($\approx 99.5\%$ similar to the input distribution). The surrogate confidence criteria is split into two cases: predicted the class label, we used the \textit{F1 Score} where \textit{F1 Score Threshold} $= 0.90$ and predicted the real label (KSTS value), we used \textit{RMSE} where \textit{RMSE Threshold} $= 0.001$. We conducted a total of $126$ experiments, averaging each experiment $20$ times, where for each average the true parameters were varied and different combinations of parameters were tested. We compared all of the strategies implemented, parametersing $1, \dots, 7$ parameters as seen in Table \ref{table1}. Each of the surrogate assisted optimisation strategies have been initialised to generate $1000$ new samples and perform three iterations every time the strategy is executed within the framework.

\subsection{Real World Experiment Setup}
The real-world dataset we used comes from the COVID-19 Data Repository by the Center for Systems Science and Engineering (CSSE) at Johns Hopkins University~\citep{covid_data}. We are interested in the number of daily infections, particularly in South Africa. However, the dataset only keeps track of the total confirmed infections per day. Therefore, we got the daily infections for the current day by getting the difference between the recorded infections for the next day and the current day (excluding the last day in the dataset). We used a seven-day moving average between 16/06/2020 -- 06/09/2020 to smooth out any local reporting anomalies. The framework configurations for this experiment are the same as mentioned above. However, the agent-based model's simulation steps have been increased to $83$ to match the number of days between the specified dates.

\subsection{Hardware Specifications}
The machine used to run our experiments consisted of an Intel Xeon CPU E5-2683 v4 @ 2.10GHz processor with 64 CPUs and 256GB of RAM using the Ubuntu 18.04.4 LTS operating system.

\section{Results and Discussion}

\begin{table*}[thb!]
\caption{Standardised $\text{L}_2$ Norm values for the optimal predicted parameter vectors using each of the strategies and surrogate models implemented.}
\label{table2}
\begin{center}
\resizebox{0.7\textwidth}{!}{
\begin{tabular}{ll|rrrrrrr}
\toprule
& & \multicolumn{7}{c}{\textbf{Standardised $\text{L}_2$ Norm}} \\ 
\textbf{Strategy} & \textbf{Surrogate} & \textbf{1} & \textbf{2} & \textbf{3} & \textbf{4} & \textbf{5} & \textbf{6} & \textbf{7}\\ 
\bottomrule\toprule
\textbf{Random}  
 & -     &  0.1778 &  \cellcolor{black!7}0.5363 &  0.7105 &  1.5249 &  2.8104 &  1.9735 &  3.1206  \\
\textbf{Sobol}   
 & -     &  0.0546 &  0.6204 &  0.9194 &  1.3720 &  2.1372 &  2.9459 &  4.6424  \\
\midrule\midrule
\multirow{3}{*}{\textbf{Random}}
 & \textbf{XGBoost} &  0.0695 &  0.6866 &  1.0335 &  0.9920 &  3.2106 &  3.2145 &  4.3614  \\
 & \textbf{DT}      &  0.0421 &  0.5888 &  0.7117 &  1.2247 &  2.5250 &  2.1988 &  4.3225  \\
 & \textbf{SVM}     &  0.0541 &  0.5378 &  0.7395 &  1.1894 &  1.8206 &  2.1682 &  3.6745  \\
\midrule
\multirow{3}{*}{\textbf{Sobol}}
 & \textbf{XGBoost} &  0.1076 &  0.7249 &  0.8582 &  1.2848 &  2.2395 &  2.8396 &  4.9996  \\
 & \textbf{DT}      &  0.1092 &  0.7215 &  0.8129 &  1.3268 &  2.0950 &  3.0756 &  4.0618  \\
 & \textbf{SVM}     &  0.0795 &  0.9444 &  1.0304 &  1.1763 &  2.1551 &  3.5414 &  3.3606  \\
\midrule\midrule
\multirow{3}{*}{\textbf{MSRS}}
 & \textbf{XGBoost} &  0.1442 
                    &  0.6426 
                    &  0.7628          
                    & \cellcolor{black!7}0.9560          
                    & 1.5134                     
                    & \cellcolor{black!14}1.6819 
                    & 3.0205  \\
 & \textbf{DT}      & 0.0463 
                    & 0.7482 
                    & \cellcolor{black!7}0.6837          
                    & 1.0738          
                    & \cellcolor{black!21}1.2975 
                    & \cellcolor{black!7} 1.8974          
                    & \cellcolor{black!7}2.8574\\
 & \textbf{SVM}     &  0.0463 
                    &  0.7436 
                    & \cellcolor{black!14}0.6654 
                    & \cellcolor{black!14}0.9479
                    & \cellcolor{black!14}1.3820 
                    & 2.6611          
                    & \cellcolor{black!21}2.3829\\
\midrule
\multirow{3}{*}{\textbf{DYCORS}}
 & \textbf{XGBoost} & 0.0679          
                    & 0.6729          
                    & 0.8196 
                    & 1.3433          
                    & 2.6203 
                    & \cellcolor{black!21}{1.6127} 
                    & 3.2048 \\
 & \textbf{DT}      & \cellcolor{black!21}{0.0281} 
                    & \cellcolor{black!14}{0.5231} 
                    & 0.8930 
                    & \cellcolor{black!21}{0.9465} 
                    & 2.6397 & 3.5522          
                    & \cellcolor{black!14}2.7680 \\
 & \textbf{SVM}     & \cellcolor{black!14}{0.0289} 
                    & 0.5827          
                    & 0.7838 & 1.2029          
                    & \cellcolor{black!7}1.4711 
                    & 2.2533          
                    & 3.9010 \\
\midrule
\multirow{3}{*}{\textbf{CMAES}}
 & \textbf{XGBoost} & 0.1175 
                    & 0.5679          
                    & 0.9984          
                    & 2.2796 
                    & 2.2526 
                    & 4.4795 
                    & 4.2173\\
 & \textbf{DT}      & 0.1423 
                    & 1.0773          
                    & 0.7987          
                    & 1.1261 
                    & 2.8731 
                    & 4.6390 
                    & 3.9230\\
 & \textbf{SVM}     & \cellcolor{black!7}0.0369 
                    & \cellcolor{black!21}{0.4593} 
                    & \cellcolor{black!21}{0.5793} 
                    & 1.0855 
                    & 3.1720 
                    & 4.5925 
                    & 3.9130\\
\bottomrule
\end{tabular}}
\end{center}

\end{table*}

\begin{table*}[thb!]
\caption{Kolmogorov-Smirnov Test Statistic (KSTS) values for the optimal predicted parameter vectors using each of the strategies and SMs implemented.}
\label{table3}
\begin{center}
\resizebox{0.7\textwidth}{!}{
\begin{tabular}{ll|rrrrrrr}
\toprule
 & \textbf{}  & \multicolumn{7}{c}{\textbf{Kolmogorov-Smirnov Test Statistic (KSTS)}}                \\ 
\textbf{Strategy} & \textbf{Surrogate} & \textbf{1} & \textbf{2} & \textbf{3} & \textbf{4} & \textbf{5} & \textbf{6} & \textbf{7}\\ 
\bottomrule\toprule
\textbf{Random}    
 & -                 & .0 &  0.0004           & 0.0018 & 0.0034 & 0.0047 &  0.0045 & 0.0043 \\
\textbf{Sobol}     
 & -                 & .0 &  \textbf{0.0000}  & 0.0035 & 0.0035 &  0.0037 & 0.0042 & 0.0039 \\
\midrule\midrule
\multirow{3}{*}{\textbf{Random}}
 & \textbf{XGBoost}  & .0 & 0.0008          & 0.0026 & 0.0034 & 0.0041 & 0.0046 & 0.0043 \\
 & \textbf{DT }      & .0 & \textbf{0.0000} & 0.0033 & 0.0028 & 0.0035 & 0.0047 & 0.0038 \\
 & \textbf{SVM}      & .0 & 0.0004          & 0.0019 & 0.0037 & 0.0044 & 0.0049 & 0.0047 \\
\midrule
\multirow{3}{*}{\textbf{Sobol}}
 & \textbf{XGBoost}  & .0 & 0.0003          & 0.0029 & 0.0034 & 0.0045 & 0.0050 & 0.0045 \\
 & \textbf{DT}       & .0 & \textbf{0.0000} & 0.0025 & 0.0027 & 0.0033 & 0.0044 & 0.0039 \\
 & \textbf{SVM}      & .0 & \textbf{0.0000} & 0.0030 & 0.0029 & 0.0041 & 0.0048 & 0.0043 \\
\midrule\midrule
\multirow{3}{*}{\textbf{MSRS}}
 & \textbf{XGBoost}  & .0 & 0.0009 & 0.0039 & 0.0034 & 0.0038 & 0.0043 & 0.0039 \\
 & \textbf{DT}       & .0 & 0.0002 & 0.0019 & 0.0032 & 0.0040 & 0.0047 & 0.0045 \\
 & \textbf{SVM}      & .0 & 0.0003 & 0.0036 & 0.0031 & 0.0045 & 0.0042 & \textbf{0.0034} \\
\midrule
\multirow{3}{*}{\textbf{DYCORS}}
 & \textbf{XGBoost} & .0 & 0.0003          & 0.0023 & 0.0031 & 0.0046          & 0.0042          & 0.0044 \\
 & \textbf{DT}      & .0 & 0.0003          & 0.0039 & 0.0030 & \textbf{0.0033} & 0.0041          & 0.0045 \\
 & \textbf{SVM}     & .0 & \textbf{0.0000} & 0.0036 & 0.0026 & 0.0041          & \textbf{0.0036} & 0.0041 \\
\midrule
\multirow{3}{*}{\textbf{CMA-ES}}
 & \textbf{XGBoost} & .0 & 0.0004 & \textbf{0.0006} & 0.0023          & 0.0051 & 0.0101 & 0.0055 \\
 & \textbf{DT}      & .0 & 0.0005 & 0.0024          & \textbf{0.0018} & 0.0034 & 0.0067 & 0.0046 \\
 & \textbf{SVM}     & .0 & 0.0005 & 0.0031          & 0.0018          & 0.0048 & 0.0109 & 0.0060 \\
\bottomrule
\end{tabular}}
\end{center}
\end{table*}

\begin{figure}[htb!]
    \centering
    \resizebox{\columnwidth}{!}{%
    \includegraphics[scale=0.5]{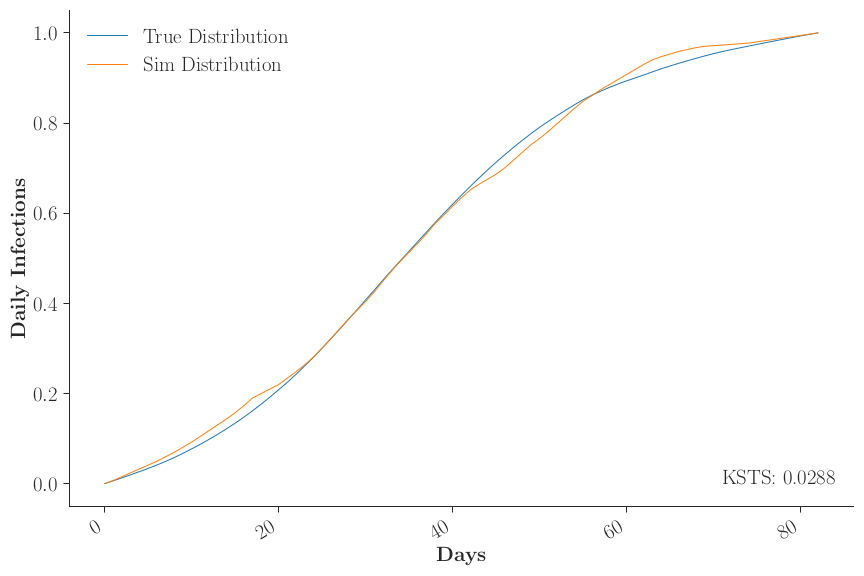}
    }
    \caption{Scaled cumulative number of infections per day between a real world infected distribution and simulated infected distribution using the DYCORS XGBoost strategy-surrogate combination.}
    \label{figreal1}
\end{figure}
\begin{figure}[htb!]
    \centering
    \resizebox{\columnwidth}{!}{%
        \includegraphics[scale=0.5]{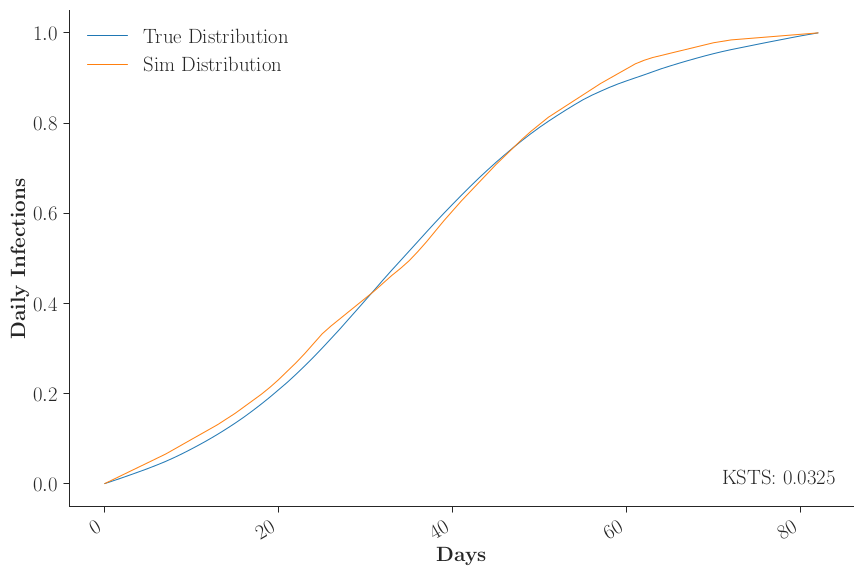}
    }
    \caption{Scaled cumulative number of infections per day between a real world infected distribution and simulated infected distribution using the MSRS SVM strategy-surrogate combination.}
    \label{figreal2}
\end{figure}

\begin{table*}[htb!]
\caption{Probability of reaching success and the speedup acquired within 98\% and 99\% of the optimal for seven parameters.}
\centering
\resizebox{0.65\textwidth}{!}{
\begin{tabular}{ll|rr|rr}
\toprule
                  &                     & \multicolumn{2}{c|}{\textbf{Success}} & \multicolumn{2}{c}{\textbf{Speedup}} \\ 
\textbf{Strategy} & \textbf{Surrogate}  & \textbf{@98\%} & \textbf{@99\%} & \textbf{@98\%} & \textbf{@99\%} \\ 
\bottomrule\toprule
\textbf{Random} &  -    &  \textbf{0.85} & \textbf{ 0.75} & 0.00 & 0.00 \\ 
\textbf{Sobol}  &  -    &  \textbf{0.85} & 0.65 & 0.00 & 0.00 \\
\midrule
\midrule
\multirow{3}{*}{\textbf{Random}}
 & \textbf{XGBoost} &  0.80          &  0.70 &  3.60 &  2.61 \\
 & \textbf{DT}      &  \textbf{0.85} & \textbf{0.75} & 3.46 &  2.61 \\
 & \textbf{SVM}     &  0.80          &  0.70 &  3.53 &  2.50 \\
\midrule
\multirow{3}{*}{\textbf{Sobol}}  
 & \textbf{XGBoost}  &  0.80          &  0.70 &  3.21 &  2.61 \\
 & \textbf{DT}       &  0.80          &  0.70 &  \textbf{4.00} & \textbf{3.10} \\
 & \textbf{SVM}      &  \textbf{0.85} &  0.70 &  3.40 &  2.40 \\ 
\midrule\midrule
\multirow{3}{*}{\textbf{MSRS}}
 & \textbf{XGBoost}  &  \textbf{0.85} &  0.70 &  \textbf{4.00} &  \textbf{2.86} \\
 & \textbf{DT}       &  \textbf{0.85} &  0.65 & 3.53 &  2.31 \\
 & \textbf{SVM}      &  0.80 &  0.70 &  2.81 &  2.65 \\
\midrule
\multirow{3}{*}{\textbf{DYCORS}} 
 & \textbf{XGBoost} &  \cellcolor{black!20}\textbf{0.85} &  \cellcolor{black!20}\textbf{0.75}  &  \cellcolor{black!20}\textbf{3.75} &  \cellcolor{black!20}\textbf{2.90} \\
 & \textbf{DT}      &  0.75          &  0.70  &  3.16 &  2.73 \\
 & \textbf{SVM}     &  \textbf{0.85} &  0.70  &  3.39 &  2.81 \\
\midrule
\multirow{3}{*}{\textbf{CMA-ES}}  
 & \textbf{XGBoost}  &  0.70          &  0.55 &  2.90 &  2.11 \\
 & \textbf{DT}       &  0.75          &  0.65 &  3.10 &  2.65 \\
 & \textbf{SVM}      &  0.75          &  0.60 &  3.16 &  2.25 \\
\bottomrule
\end{tabular}}

\label{table4}
\end{table*}

\begin{table}[htb!]
\centering
\caption{
Predicted optimal parameter vectors for parameterising seven parameters for the DYCORS XGBoost and MSRS SVM strategy-surrogate combinations. 
}\resizebox{1\columnwidth}{!}{
\begin{tabular}{l|l|l}
\toprule
\multicolumn{1}{c}{\textbf{Parameters}} & \multicolumn{1}{|c}{\textbf{DYCORS XGBoost}} & \multicolumn{1}{|c}{\textbf{MSRS SVM}}\\
\bottomrule \toprule
1. Transmission Probability ($\beta$)   &   0.133    &     0.142    \\ 
2. Reinfection Probability              &   0.469    &     0.664    \\ 
3. Death Probability                    &   0.072    &     0.024    \\ 
4. Infection Period                     &   18.0     &     26.0     \\ 
5. Detection Time                       &   11.0     &     24.0     \\ 
6. Speed                                &   0.019    &     0.004    \\ 
7. Interaction Radius                   &   0.007    &     0.006    \\ 
\bottomrule
\end{tabular}}
\label{table5}
\end{table}

In the following tables and figures, we present the results of the above experiment, followed by a discussion of the results. The Standardised L$_{2}$ Norm values show the distance between the true (synthetic) parameter vector and the ABMS Framework's optimal prediction. The optimal prediction is the parameter vector (estimate) with the lowest Kolmogorov-Smirnov Test Statistic (KSTS) value. The sanity check assesses whether our ABMS Framework and a strategy-surrogate combination can approximate the parameters that generated the actual distribution. The values presented in Table \ref{table2} are relatively close to zero, which implies our sanity check holds. Also,  we have highlighted the top three lowest Standardised L$_{2}$ Norm values for each of the strategy-surrogate combinations considered. The results show that overall the MSRS strategy can best approximate the synthetic parameters. In particular, MSRS is relatively good at attaining the synthetic parameters for three or more parameters, whereas DYCORS and CMA-ES are relatively good for $1$--$3$ parameters. The values presented in both Table \ref{table2} and \ref{table3} show us that the Support Vector Machine (SVM) and Decision Tree (DT) surrogate gets a more significant majority of optimal values (i.e. the lowest values), whereas this is not the case for XGBoost.

The KSTS values tell us how close we can get to the actual (synthetic) infected data distribution. Table~\ref{table3} shows that for one and two parameters, we are still in a relatively low dimensional space, and as such, the baselines and the surrogate assisted samplers perform the best. As we increase the dimensionality of the parameter space, the CMA-ES strategy is optimal for three and four parameters. When moving towards a higher dimensional space of five or six parameters, the DYCORS strategy attains the lowest KSTS value, and for seven parameters, MSRS has the lowest value. Inspecting Table \ref{table3} more closely, we note that all techniques can reasonably approximate the synthetic distribution and that no strategy-surrogate combination stands out clearly from the others. Also, the KSTS values presented in Table \ref{table3} are significant regarding the p-values of the Kolmogorov-Smirnov test. 

When parameterising all seven parameters, we see that MSRS SVM can approximate the synthetic parameters the best with the lowest Standardised L$_{2}$ Norm value (2.3829) on average. Also, MSRS SVM can get the lowest KSTS value (0.0034). This result implies that MSRS SVM can predict a parameter vector that generates a simulated infection distribution when run through the ABM, most similar to the synthetic infection distribution.  

The probabilities of reaching an optimal solution (success) within $98$\% and $99$\% of the optimal value are captured in Table \ref{table4}. Success is defined as a measure of similarity to the synthetic data distribution. For example, success at $98$\% implies that a method can produce a simulated distribution that is $98$\% similar to the synthetic distribution that we try to replicate. In the same table, we also show the speedup attained by each strategy-surrogate combination. Across all implemented strategy-surrogate combinations, we obtain a significant speedup compared to the baselines.

Table \ref{table4} shows that Random, Random DT and DYCORS XGBoost have the highest probability of reaching success at $98$\% and $99$\% in that specific order. Also, the Sobol DT, MSRS XGBoost, and DYCORS XGBoost strategy-surrogate combinations achieve the most speedup. Although Random attains a high probability of reaching success, it is not efficient in that it provides no speedup compared to the other methods. The DYCORS XGBoost strategy-surrogate combination is optimal when considering both probabilities of reaching success and speedup. One of the limitations is that we can only get $99$\% accuracy with a $75$\% probability. This suboptimal probability means that we would need to run the model more than once to ensure the correct outcome. Running more than once would negate some of the speedup achieved. 

In general, we would prefer utilising a strategy-surrogate combination that can predict an optimal parameter vector that perfectly matches the actual (real) data distribution in terms of accuracy and speedup (DYCORS XGBoost). However, we may also want to utilise a strategy-surrogate combination that gets as close as possible to the true
(real) parameter vector, generating the actual data distribution (MSRS SVM). In Figures \ref{figreal1} and \ref{figreal2}, we show a comparison between the actual cumulative daily infection distribution and the simulated distribution obtained by the ABMS Framework using DYCORS XGBoost and MSRS SVM, respectively. The simulated distribution is generated by taking the optimal prediction from both strategy-surrogate combinations and evaluate them through the ABM. We fail to reject the Kolmogorov-Smirnov Test's null hypothesis, as both optimal predictions generated distributions that closely model the real-world distribution. MSRS SVM can get a KSTS value of $0.0325 \approx 96.75$\%, similar to the actual distribution. DYCORS XGBoost can get a KSTS value of $0.0288 \approx 97.12$\%, similar to the actual distribution. The optimal predicted parameters for each strategy-surrogate combination is shown in Table \ref{table5}. The infection period and detection time values attained by DYCORS XGBoost are pretty similar to the virus characteristics of COVID-19\footnote{https://www.health.harvard.edu/diseases-and-conditions/if-youve-been-exposed-to-the-coronavirus}.







\section{Conclusion}
We have implemented a more extensive and adaptive agent-based modelling and simulation (ABMS) Framework. Our framework can effectively swap out parameterisation strategies and surrogate models (SMs) to parameterise infectious disease agent-based models (ABMs). We show that in terms of the lowest Kolmogorov-Smirnov Test Statistic (KSTS) values, we achieve better than parity across all parameters compared to the surrogate assisted sampling strategies and the baselines. The Decision Tree (DT) and the Support Vector Machine (SVM) surrogates are on par with each other to attain the lowest KSTS values overall, whereas XGBoost is not reliable. Using the Standardised L$_2$ Norm values, we show that MSRS SVM is the best strategy-surrogate combination overall in terms of getting closest to the true synthetic parameters. One of the significant challenges in ABMS is the time required to parameterise an ABM is cumbersome. We have shown that DYCORS XGBoost attains the highest probability of replicating the actual (synthetic) data distribution within $98$\% and $99$\% success and achieves the most considerable speedup in comparison to the baselines and evaluated strategy-surrogate pairs. DYCORS XGBoost and MSRS SVM were both used to parameterise the infectious disease ABM within our ABMS Framework. DYCORS XGBoost and MSRS SVM attained an optimal parameter vector that generated cumulative daily infection distributions that are $97.12$\% and $96.75$\% similar to the actual (real) distribution, respectively. The real distribution represents the cumulative number of daily infections between the dates $16/06/2020$--$06/09/2020$ in South Africa. Moreover, we observe the DYCORS XGBoost predicts a parameter vector representing the characteristics of COVID-19 better than MSRS SVM. Our future work aims to improve the ABM's complexity and evaluate different types of ABMs within our framework to assess overall robustness.

%
\section*{Conflict of interest}
The financial assistance of the National Research Foundation (NRF) towards this research is hereby acknowledged. Opinions expressed and conclusions arrived at, are those of the author and are not necessarily to be attributed to the NRF.

\bibliographystyle{plainnat}
\bibliography{bib}   

%
%

\end{document}